\RequirePackage{fix-cm}
\documentclass[twocolumn]{svjour3}          
\smartqed  

\usepackage{graphicx}
\usepackage{cite}
\usepackage{authblk}
\usepackage{tikz}
\usepackage{amsfonts}
\usepackage{hhline}
\usepackage{amsmath}
\usepackage{mathtools}
\usepackage{url}
\def\checkmark{\tikz\fill[scale=0.4](0,.35) -- (.25,0) -- (1,.7) -- (.25,.15) -- cycle;} 

\DeclarePairedDelimiter\floor{\lfloor}{\rfloor}
\newcommand{\norm}[1]{\| #1 \|}

\begin{document}

\title{EfficientPose: Scalable single-person pose estimation}
\titlerunning{EfficientPose}   

\author{Daniel Groos\textsuperscript{1} \and Heri Ramampiaro\textsuperscript{2} \and Espen A. F. Ihlen\textsuperscript{1}}
\authorrunning{Groos et al.}

\institute{
    Daniel Groos\\
    \quad \quad daniel.groos@ntnu.no\\\\
    Heri Ramampiaro\\
    \quad \quad heri@ntnu.no \\\\
    Espen A. F. Ihlen\\
    \quad \quad espen.ihlen@ntnu.no \\\\
\textsuperscript{1} Department of Neuromedicine and Movement Science, Norwegian University of Science and Technology, Trondheim, Norway\\\\
\textsuperscript{2} Department of Computer Science, Norwegian University of Science and Technology, Trondheim, Norway\\
}


    \date{Received: date / Accepted: date}
    \maketitle
        \begin{abstract}
        \hfill\break
        Single-person human pose estimation facilitates markerless movement analysis in sports, as well as in clinical applications. Still, state-of-the-art models for human pose estimation generally do not meet the requirements of real-life applications. The proliferation of deep learning techniques has resulted in the development of many advanced approaches. However, with the progresses in the field, more complex and inefficient models have also been introduced, which have caused tremendous increases in computational demands. To cope with these complexity and inefficiency challenges, we propose a novel convolutional neural network architecture, called EfficientPose, which exploits recently proposed EfficientNets in order to deliver efficient and scalable single-person pose estimation. EfficientPose is a family of models harnessing an effective multi-scale feature extractor and computationally efficient detection blocks using mobile inverted bottleneck convolutions, while at the same time ensuring that the precision of the pose configurations is still improved. Due to its low complexity and efficiency, EfficientPose enables real-world applications on edge devices by limiting the memory footprint and computational cost. The results from our experiments, using the challenging MPII single-person benchmark, show that the proposed EfficientPose models substantially outperform the widely-used OpenPose model both in terms of accuracy and computational efficiency. In particular, our top-performing model achieves state-of-the-art accuracy on single-person MPII, with low-complexity ConvNets.
        \end{abstract}

\section{Introduction}
\label{intro}

Single-person human pose estimation (HPE) refers to the computer vision task of localizing human skeletal keypoints of a person from an image or video frames. Single-person HPE has many real-world applications, ranging from outdoor activity recognition and computer animation to clinical assessments of motor repertoire and skill practice among professional athletes. The proliferation of deep convolutional neural networks (ConvNets) has advanced HPE and further widen its application areas. ConvNet-based HPE with its increasingly complex network structures, combined with transfer learning, is a very challenging task. However, the availability of high-performing ImageNet~\cite{deng2009imagenet} backbones, together with large tailor-made datasets, such as MPII for 2D pose estimation~\cite{andriluka14cvpr}, has facilitated the development of new improved methods to address the challenges.

An increasing trend in computer vision has driven towards more efficient models~\cite{sandler2018mobilenetv2, tan2019mnasnet, elsen2019fast}. Recently, EfficientNet~\cite{tan2019efficientnet} was released as a scalable ConvNet architecture, setting benchmark record on ImageNet with a more computationally efficient architecture. However, within human pose estimation, there is still a lack of architectures that are both accurate and computationally efficient at the same time. In general, current state-of-the-art architectures are computationally expensive and highly complex, thus making them hard to replicate, cumbersome to optimize, and impractical to embed into real-world applications. 

The OpenPose network~\cite{cao2018openpose} (OpenPose for short) has been one of the most applied HPE methods in real-world applications. It is also the first open-source real-time system for HPE. OpenPose was originally developed for multi-person HPE, but has in recent years been frequently applied to various single-person applications within clinical research and sport sciences~\cite{nakai2018prediction, noori2019robust, firdaus2019recognizing}. The main drawback with OpenPose is that the level of detail in keypoint estimates is limited due to its low-resolution outputs. This makes OpenPose less suitable for precision-demanding applications, such as elite sports and medical assessments, which all depend on high degree of precision in the assessment of movement kinematics. Moreover, by spending $160$ billion floating-point operations (GFLOPs) per inference, OpenPose is considered highly inefficient. Despite these issues, OpenPose seems to remain a commonly applied network for single-person HPE performing markerless motion capture from which critical decisions are based upon~\cite{vitali2019new, barra2019gait}. 

In this paper, we stress the lack of publicly available methods for single-person HPE that are both computationally efficient and effective in terms of estimation precision. To this end, we exploit recent advances in ConvNets and propose an improved approach called EfficientPose. Our main idea is to modify OpenPose into a family of scalable ConvNets for high-precision and computationally efficient single-person pose estimation from 2D images. To assess the performance of our approach, we perform two separate comparative studies. First, we evaluate the EfficientPose model by comparing it against the original OpenPose model on single-person HPE. Second, we compare it against the current state-of-the-art single-person HPE methods on the official MPII challenge, focusing on accuracy as a function of the number of parameters. The proposed EfficientPose models aim to elicit high computational efficiency, while bridging the gap in availability of high-precision HPE networks.

In summary, the main contributions of this paper are the following: 
\begin{itemize}
 \item We propose an improvement of OpenPose, called EfficientPose, that can overcome the shortcomings of the popular OpenPose network on single-person HPE with improved level of precision, rapid convergence during optimization, low number of parameters, and low computational cost.
 \item With EfficientPose, we suggest an approach providing scalable models that can suit various demands, enabling a trade-off between accuracy and efficiency across diverse application constraints and limited computational budgets.
 \item We propose a new way to incorporate mobile ConvNet components, which can address the need for computationally efficient architectures for HPE, thus facilitating real-time HPE on the edge.
 \item We perform an extensive comparative study to evaluate our approach. Our experimental results show that the proposed method achieves significantly higher efficiency and accuracy in comparison to the baseline method, OpenPose. In addition, compared to existing state-of-the-art methods, it achieves competitive results, with a much smaller number of parameters.
\end{itemize}

The remainder of this paper is organized as follows: Section~\ref{sec:related} describes the architecture of OpenPose and highlights research which it can be improved from. Based on this, Section~\ref{sec:efficientpose} presents our proposed ConvNet-based approach, EfficientPose. Section~\ref{sec:experiments} describes our experiments and presents the results from comparing EfficientPose with OpenPose and other existing approaches. Section~\ref{sec:discussion} discusses our findings and suggests potential future studies. Finally, Section~\ref{sec:conclusion} summarizes and concludes the paper.

For the sake of reproducibility, we will make the EfficientPose models available at \url{https://github.com/daniegr/EfficientPose}.

\section{Related work}
\label{sec:related}

\begin{figure*}
\begin{center}
\includegraphics[width=\textwidth]{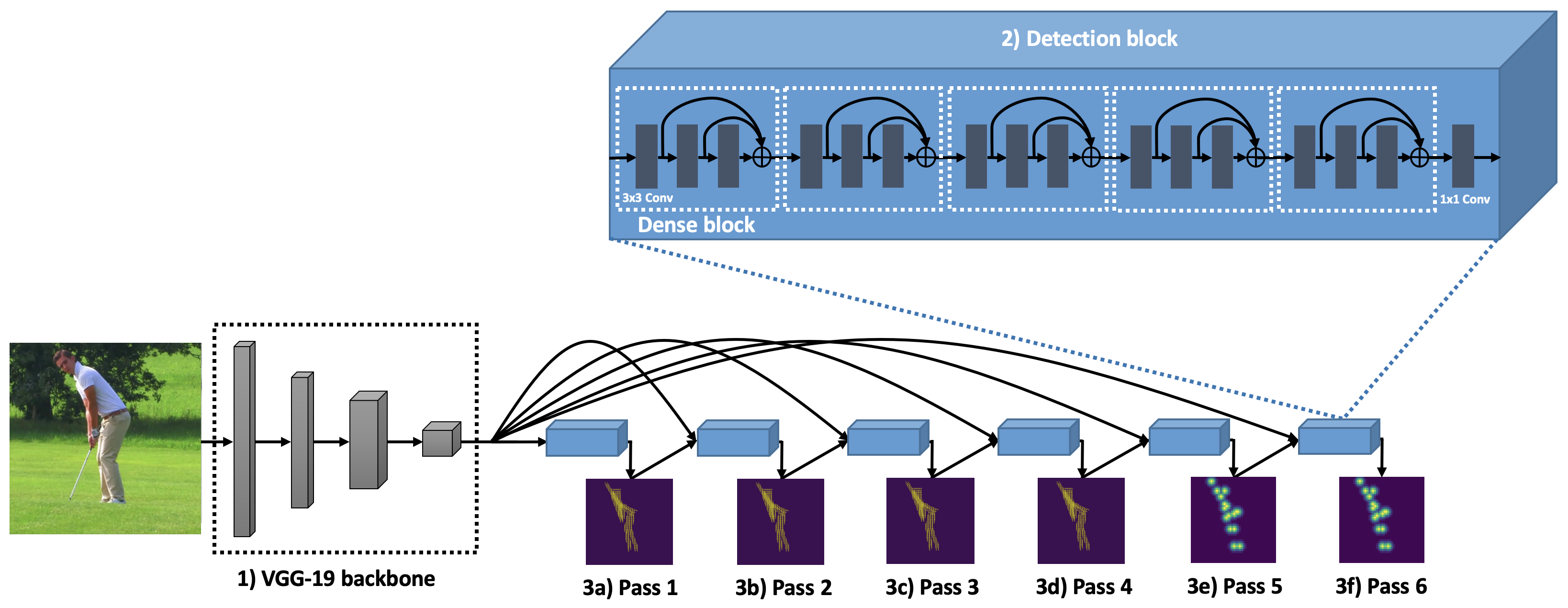}
\caption{OpenPose architecture utilizing 1) VGG-19 feature extractor, and 2) 4+2 passes of detection blocks performing 4+2 passes of estimating part affinity fields (3a-d) and confidence maps (3e and 3f)}
\label{fig:openpose}       
\end{center}
\end{figure*}

The proliferation of ConvNets for HPE following the success of DeepPose~\cite{toshev2014deeppose} has set the path for accurate HPE. With OpenPose, Cao et al.~\cite{cao2018openpose} made HPE available to the public. As depicted by Figure~\ref{fig:openpose}, OpenPose comprises a multi-stage architecture performing a series of detection passes. Provided an input image of $368 \times 368$ pixels, OpenPose utilizes an ImageNet pretrained VGG-19 backbone~\cite{simonyan2014very} to extract basic features (step 1 in Figure~\ref{fig:openpose}). The features are supplied to a DenseNet-inspired detection block (step 2) arranged as five dense blocks~\cite{huang2017densely}, each containing three $3 \times 3$ convolutions with PReLU activations~\cite{he2015delving}. The detection blocks are stacked in a sequence. First, four passes (step 3a-d in Figure~\ref{fig:openpose}) of part affinity fields~\cite{cao2017realtime} map the associations between body keypoints. Subsequently, two detection passes (step 3e and 3f) predict keypoint heatmaps~\cite{tompson2014joint} to obtain refined keypoint coordinate estimates. In terms of level of detail in the keypoint coordinates, OpenPose is restricted by its output resolution of $46 \times 46$ pixels.

The OpenPose architecture can be improved by recent advancements in ConvNets, as follows: First, automated network architecture search has found backbones~\cite{tan2019efficientnet, zoph2018learning, tan2019mixconv} that are more precise and efficient in image classification than VGG and ResNets~\cite{simonyan2014very, he2016deep}. In particular, Tan and Le~\cite{tan2019efficientnet} proposed compound model scaling to balance the image resolution, width (number of network channels), and depth (number of network layers). This resulted in scalable convolutional neural networks, called EfficientNets~\cite{tan2019efficientnet}, with which the main goal was to provide lightweight models with a sensible trade-off between model complexity and accuracy across various computational budgets. For each model variant EfficientNet-B$\phi$, from the least computationally expensive one being EfficientNet-B0 to the most accurate model, EfficientNet-B7 ($\phi \in [0,7] \in \mathbb{Z}^{\geq}$), the total number of FLOPs increases by a factor of $2$, given by
\begin{equation}
(\alpha \cdot \beta^2 \cdot \gamma^2)^{\phi} \approx 2^{\phi}.
\label{eq:scaling}
\end{equation}
Here, $\alpha$, $\beta$ and $\gamma$ denote the coefficients for depth, width, and resolution, respectively, and are set as
\begin{equation}
\alpha=1.2, \beta=1.1, \gamma=1.15.
\label{eq:coefficients}
\end{equation}
Second, parallel multi-scale feature extraction has improved the precision levels in HPE~\cite{newell2016stacked, ke2018multi, sun2019deep, yang2017learning}, emphasizing both high spatial resolution and low-scale semantics. However, existing multi-scale approaches in HPE are computationally expensive, both due to their large size and high computational requirements. For example, a typical multi-scale HPE approach has often a size of $16 - 58$ million parameters and requires $10 - 128$ GFLOPS~\cite{sun2019deep, chu2017multi, zhang2019human, tang2018deeply, yang2017learning, newell2016stacked, rafi2016efficient}. To cope with this, we propose cross-resolution features, operating on high- and low-resolution input images, to integrate features from multiple abstraction levels with low overhead in network complexity and with high computational efficiency. Existing works on Siamese ConvNets have been promising in utilizing parallel network backbones~\cite{gao2019siamese, gao2020learning}. 
Third, mobile inverted bottleneck convolution (MBConv)~\cite{sandler2018mobilenetv2} with built-in squeeze-and-excitation (SE)~\cite{hu2018squeeze} and Swish activation~\cite{DBLP:conf/iclr/RamachandranZL18} integrated in EfficientNets has proven more accurate in image classification tasks~\cite{tan2019efficientnet, tan2019mixconv} than regular convolutions~\cite{he2016deep, huang2017densely, szegedy2017inception}, while substantially reducing the computational costs~\cite{tan2019efficientnet}. The efficiency of MBConv modules stem from the depthwise convolutions operating in a channel-wise manner~\cite{sifre2014rigid}. With this approach, it is possible to reduce the computational cost by a factor proportional to the number of channels~\cite{tan2019mixconv}.
Hence, by replacing the regular $3 \times 3$ convolutions with up to $384$ input channels in the detection blocks of OpenPose with MBConvs, we can obtain more computationally efficient detection blocks. Further, SE selectively emphasizes discriminative image features~\cite{hu2018squeeze}, which may reduce the required number of convolutions and detection passes by providing a global perspective on the estimation task at all times. Using MBConv with SE may have the potential to decrease the number of dense blocks in OpenPose. Fourth, transposed convolutions with bilinear kernel~\cite{long2015fully} scale up the low-resolution feature maps, thus enabling a higher level of detail in the output confidence maps. 

By building upon the work of Tan and Le~\cite{tan2019efficientnet}, we present a pool of scalable models for single-person HPE that is able to overcome the shortcomings of the commonly adopted OpenPose architecture. This enables trading off between accuracy and efficiency across different computational budgets in real-world applications. The main advantage of this is that we can use ConvNets that are small and computationally efficient enough to run on edge devices with little memory and low processing power, which is impossible with OpenPose. 

\section{The EfficientPose approach}
\label{sec:efficientpose}

In this section, we explain in details the EfficientPose approach. This includes a detailed description of the EfficientPose architecture in light of the OpenPose architecture, and a brief introduction to the proposed variants of EfficientPose.

\subsection{Architecture}
\label{sec:architecture}
\begin{figure*}
\begin{center}
\includegraphics[width=\textwidth]{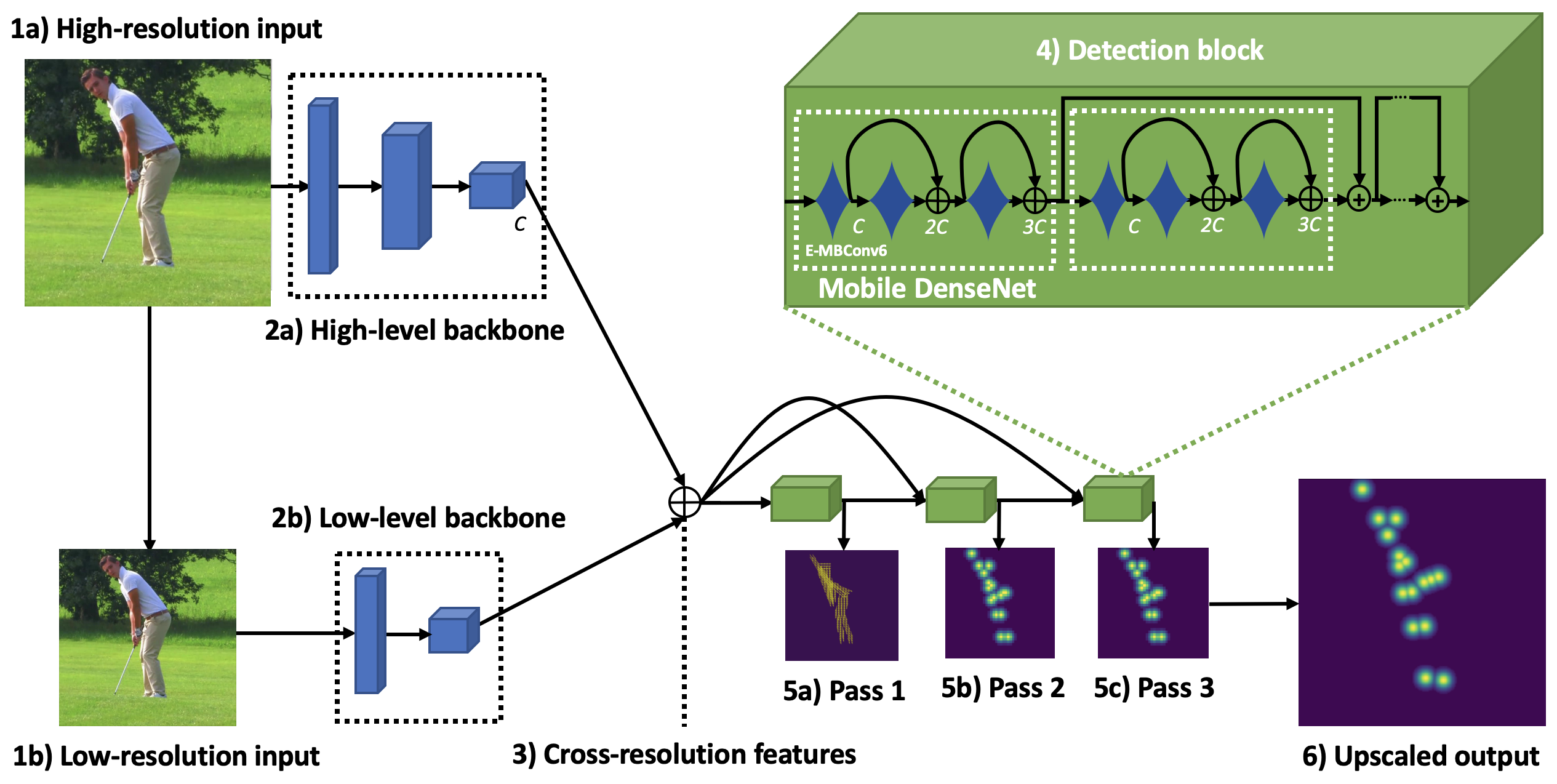}
\caption{Proposed architecture comprising 1a) high-resolution and 1b) low-resolution inputs, 2a) high-level and 2b) low-level EfficientNet backbones combined into 3) cross-resolution features, 4) Mobile DenseNet detection blocks, 1+2 passes for estimation of part affinity fields (5a) and confidence maps (5b and 5c), and 6) bilinear upscaling}
\label{fig:efficientpose}     
\end{center}
\end{figure*}

Figure~\ref{fig:openpose} and Figure~\ref{fig:efficientpose} depict the architectures of OpenPose and EfficientPose, respectively. As can be observed in these two figures, although being based on OpenPose, the EfficientPose architecture is different from the OpenPose architecture in several aspects, including 1) both high and low-resolution input images, 2) scalable EfficientNet backbones, 3) cross-resolution features, 4) and 5) scalable Mobile DenseNet detection blocks in fewer detection passes, and 6) bilinear upscaling. For a more thorough component analysis of EfficientPose, see Appendix~\ref{sec:ablation}.

The input of the network consists of high and low-resolution images (1a and 1b in Figure~\ref{fig:efficientpose}). To get the low-resolution image, the high-resolution image is downsampled into half of its pixel height and width, through an initial average pooling layer. 

\begin{figure*}
\begin{center}
\includegraphics[width=\textwidth]{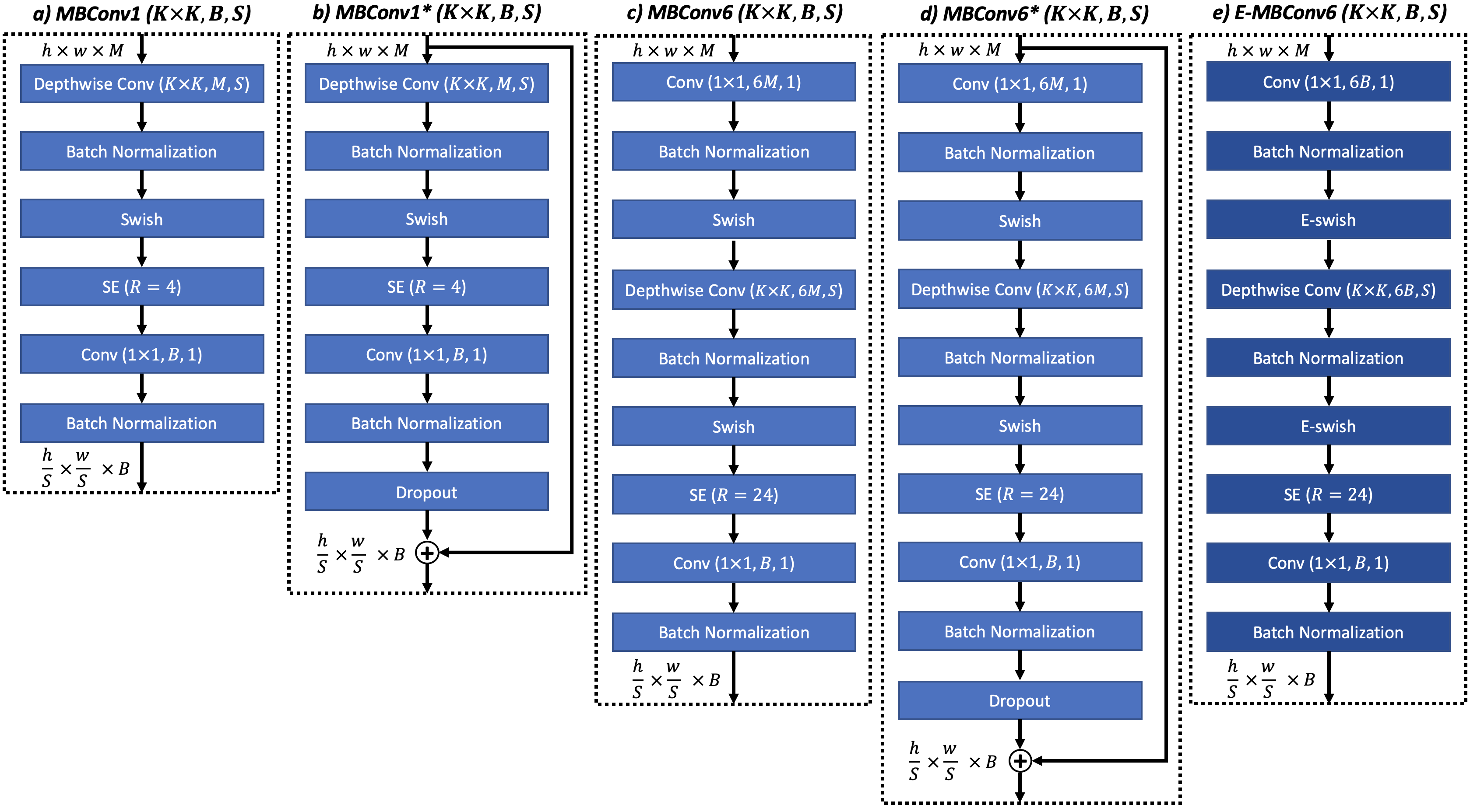}
\caption{The composition of MBConvs. From left: a-d) $MBConv (K \times K, B, S)$ in EfficientNets performs depthwise convolution with filter size $K \times K$ and stride $S$, and outputs $B$ feature maps. $MBConv^{*}$ (b and d) extends regular MBConvs by including dropout layer and skip connection. e) $E{\text -}MBConv6 (K \times K, B, S)$ in Mobile DenseNets adjusts $MBConv6$ with E-swish activation and number of feature maps in expansion phase as $6B$. All MBConvs take as input $M$ feature maps with spatial height and width of $h$ and $w$, respectively. $R$ is the reduction ratio of SE}
\label{fig:mbconvs}
\end{center}
\end{figure*}

The feature extractor of EfficientPose is composed of the initial blocks of EfficientNets~\cite{tan2019efficientnet} pretrained on ImageNet (step 2a and 2b in Figure~\ref{fig:efficientpose}). High-level semantic information is obtained from the high-resolution image using the initial three blocks of a high-scale EfficientNet with $\phi \in [2,7]$ (see Equation~\ref{eq:scaling}), outputting C feature maps (2a in Figure~\ref{fig:efficientpose}). Low-level local information is extracted from the low-resolution image by the first two blocks of a lower-scale EfficientNet-backbone (2b in Figure~\ref{fig:efficientpose}) in the range $\phi \in [0,3]$. Table~\ref{tab:efficientnets} provides an overview of the composition of EfficientNet backbones, from low-scale B0 to high-scale B7. The first block of EfficientNets utilizes the MBConvs shown in Figure~\ref{fig:mbconvs}a and~\ref{fig:mbconvs}b, whereas the second and third blocks comprise the MBConv layers in Figure~\ref{fig:mbconvs}c and~\ref{fig:mbconvs}d. 

\begin{table*}[htbp]
\centering
\small
\setlength\tabcolsep{2pt}
\caption{The architecture of the initial three blocks of relevant EfficientNet backbones. For $Conv (K \times K, N, S)$, $K \times K$ denotes filter size, $N$ is number of output feature maps, and $S$ is stride. $BN$ denotes batch normalization. $I$ defines input size, corresponding with image resolution on ImageNet, whereas $\alpha^{\phi}$ refers to the depth factor as determined by Equation~\ref{eq:scaling}}
\label{tab:efficientnets} 
\begin{tabular}{|c|c|c|c|c|c|c|c|}
\hline
\textbf{Block}                                                 & \textbf{B0}                                                     & \textbf{B1}                                                             & \textbf{B2}                    & \textbf{B3}                                                             & \textbf{B4}                                                             & \textbf{B5}                                                             & \textbf{B7}                                                             \\ \hline
$1$                                             & \multicolumn{3}{c|}{\begin{tabular}[c]{@{}c@{}}$Conv(3 \times 3, 32, 2)$\\ $BN$\\ $Swish$\end{tabular}}                                                                                 & \begin{tabular}[c]{@{}c@{}}$Conv(3 \times 3, 40, 2)$\\ $BN$\\ $Swish$\end{tabular}   & \multicolumn{2}{c|}{\begin{tabular}[c]{@{}c@{}}$Conv(3 \times 3, 48, 2)$\\ $BN$\\ $Swish$\end{tabular}}                                                        & \begin{tabular}[c]{@{}c@{}}$Conv(3 \times 3, 64, 2)$\\ $BN$\\ $Swish$\end{tabular}   \\ \cline{2-8} 
                                                               & \multicolumn{3}{c|}{\begin{tabular}[c]{@{}c@{}}$MBConv1$\\ $(3 \times 3, 16, 1)$\end{tabular}}                                                                                        & \multicolumn{3}{c|}{\begin{tabular}[c]{@{}c@{}}$MBConv1$\\ $(3 \times 3, 24, 1)$\end{tabular}}                                                                                                                                         & \begin{tabular}[c]{@{}c@{}}$MBConv1$\\ $(3 \times 3, 32, 1)$\end{tabular}          \\ \cline{2-8} 
                                                               & $-$                                                               & \multicolumn{2}{c|}{\begin{tabular}[c]{@{}c@{}}$MBConv1^{*}$\\ $(3 \times 3, 16, 1)$\end{tabular}}                     & \multicolumn{2}{c|}{\begin{tabular}[c]{@{}c@{}}$MBConv1^{*}$\\ $(3 \times 3, 24, 1)$\end{tabular}}                                                              & \begin{tabular}[c]{@{}c@{}}$\begin{bmatrix} 
  MBConv1^{*}\\ 
  (3 \times 3, 24, 1) 
\end{bmatrix} \times 2$
                                                               \end{tabular} & \begin{tabular}[c]{@{}c@{}}$\begin{bmatrix} 
  MBConv1^{*}\\ 
  (3 \times 3, 32, 1) 
\end{bmatrix} \times 3$\end{tabular} \\ \hline
$2$                                             & \multicolumn{3}{c|}{\begin{tabular}[c]{@{}c@{}}$MBConv6$\\ $(3 \times 3, 24, 2)$\end{tabular}}                                                                                        & \multicolumn{2}{c|}{\begin{tabular}[c]{@{}c@{}}$MBConv6$\\ $(3 \times 3, 32, 2)$\end{tabular}}                                                               & \begin{tabular}[c]{@{}c@{}}$MBConv6$\\ $(3 \times 3, 40, 2)$\end{tabular}          & \begin{tabular}[c]{@{}c@{}}$MBConv6$\\ $(3 \times 3, 48, 2)$\end{tabular}          \\ \cline{2-8} 
                                                               & \begin{tabular}[c]{@{}c@{}}$MBConv6^{*}$\\ $(3 \times 3, 24, 1)$\end{tabular} & \multicolumn{2}{c|}{\begin{tabular}[c]{@{}c@{}}$\begin{bmatrix} 
  MBConv6^{*}\\ 
  (3 \times 3, 24, 1) 
\end{bmatrix} \times 2$\end{tabular}}             & \begin{tabular}[c]{@{}c@{}}$\begin{bmatrix} 
  MBConv6^{*}\\ 
  (3 \times 3, 32, 1) 
\end{bmatrix} \times 2$\end{tabular} & \begin{tabular}[c]{@{}c@{}}$\begin{bmatrix} 
  MBConv6^{*}\\ 
  (3 \times 3, 32, 1) 
\end{bmatrix} \times 3$\end{tabular} & \begin{tabular}[c]{@{}c@{}}$\begin{bmatrix} 
  MBConv6^{*}\\ 
  (3 \times 3, 40, 1) 
\end{bmatrix} \times 4$\end{tabular} & \begin{tabular}[c]{@{}c@{}}$\begin{bmatrix} 
  MBConv6^{*}\\ 
  (3 \times 3, 48, 1) 
\end{bmatrix} \times 6$\end{tabular} \\ \hline
$3$                                             & \multicolumn{2}{c|}{\begin{tabular}[c]{@{}c@{}}$MBConv6$\\ $(5 \times 5, 40, 2)$\end{tabular}}                                                       & \multicolumn{2}{c|}{\begin{tabular}[c]{@{}c@{}}$MBConv6$\\ $(5 \times 5, 48, 2)$\end{tabular}}                      & \begin{tabular}[c]{@{}c@{}}$MBConv6$\\ $(5 \times 5, 56, 2)$\end{tabular}          & \begin{tabular}[c]{@{}c@{}}$MBConv6$\\ $(5 \times 5, 64, 2)$\end{tabular}          & \begin{tabular}[c]{@{}c@{}}$MBConv6$\\ $(5 \times 5, 80, 2)$\end{tabular}          \\ \cline{2-8} 
                                                               & \begin{tabular}[c]{@{}c@{}}$MBConv6^{*}$\\ $(5 \times 5, 40, 1)$\end{tabular} & \begin{tabular}[c]{@{}c@{}}$\begin{bmatrix} 
  MBConv6^{*}\\ 
  (5 \times 5, 40, 1) 
\end{bmatrix} \times 2$\end{tabular} & \multicolumn{2}{c|}{\begin{tabular}[c]{@{}c@{}}$\begin{bmatrix} 
  MBConv6^{*}\\ 
  (5 \times 5, 48, 1) 
\end{bmatrix} \times 2$\end{tabular}}             & \begin{tabular}[c]{@{}c@{}}$\begin{bmatrix} 
  MBConv6^{*}\\ 
  (5 \times 5, 56, 1) 
\end{bmatrix} \times 3$\end{tabular} & \begin{tabular}[c]{@{}c@{}}$\begin{bmatrix} 
  MBConv6^{*}\\ 
  (5 \times 5, 64, 1) 
\end{bmatrix} \times 4$\end{tabular} & \begin{tabular}[c]{@{}c@{}}$\begin{bmatrix} 
  MBConv6^{*}\\ 
  (5 \times 5, 80, 1) 
\end{bmatrix} \times 6$\end{tabular} \\ \hhline{|=|=|=|=|=|=|=|=|}
$I$                                                     & $224 \times 224$                                                       & $240 \times 240$                                                               & $260 \times 260$                      & $300 \times 300$                                                               & $380 \times 380$                                                               & $456 \times 456$                                                               & $600 \times 600$                                                               \\ \hline
$C$                                                     & \multicolumn{2}{c|}{$40$}                                                                                                                   & \multicolumn{2}{c|}{$48$}                                                                                  & $56$                                                                      & $64$                                                                      & $80$                                                                      \\ \hline
$\alpha^{\phi}$ & $1.2^{0}=1.0$                                  & $1.2^{1}=1.2$                                           & $1.2^{2}=1.4$  & $1.2^{3}=1.7$                                           & $1.2^{4}=2.1$                                           & $1.2^{5}=2.5$                                          & $1.2^{7}=3.6$                                           \\ \hline
\end{tabular}
\end{table*}

The features generated by the low-level and high-level EfficientNet backbones are concatenated to yield cross-resolution features (step 3 in Figure~\ref{fig:efficientpose}). This enables the EfficientPose architecture to selectively emphasize important local factors from the image of interest and the overall structures that guide high-quality pose estimation. In this way, we enable an alternative simultaneous handling of different features at multiple abstraction levels. 

From the extracted features, the desired keypoints are localized through an iterative detection process, where each detection pass performs supervised prediction of output maps. Each detection pass comprises a detection block and a single $1 \times 1$ convolution for output prediction. The detection blocks across all detection passes elicit the same basic architecture, comprising Mobile DenseNets (see step 4 in Figure~\ref{fig:efficientpose}). Data from Mobile DenseNets are forwarded to subsequent layers of the detection block using residual connections. The Mobile DenseNet is inspired by DenseNets~\cite{huang2017densely} supporting reuse of features, avoiding redundant layers, and MBConv with SE, thus enabling low memory footprint. In our adaptation of the MBConv operation ($E{\text -}MBConv6\\(K \times K, B, S)$ in Figure~\ref{fig:mbconvs}e), we consistently utilize the highest performing combination from~\cite{tan2019mnasnet}, i.e., a kernel size ($K \times K$) of $5 \times 5$ and an expansion ratio of $6$. We also avoid downsampling (i.e., $S = 1$) and scale the width of Mobile DenseNets by outputting number of channels relative to the high-level backbone ($B = C$). We modify the original $MBConv6$ operation by incorporating E-swish as activation function with $\beta$ value of $1.25$~\cite{gagana2018activation}. This has a tendency to accelerate progression during training compared to the regular Swish activation~\cite{DBLP:conf/iclr/RamachandranZL18}. We also adjust the first $1 \times 1$ convolution to generate a number of feature maps relative to the output feature maps $B$ rather than the input channels $M$. This reduces the memory consumption and computational latency since $B \leq M$, with $C \leq M \leq 3C$. With each Mobile DenseNet consisting of three consecutive $E{\text -}MBConv6$ operations, the module outputs $3C$ feature maps. 

EfficientPose performs detection in two rounds (step 5a-c in Figure~\ref{fig:efficientpose}). First, the overall pose of the person is anticipated through a single pass of skeleton estimation (5a). This aims to facilitate the detection of feasible poses and to avoid confusion in case of several persons being present in an image. Skeleton estimation is performed utilizing part affinity fields as proposed in~\cite{cao2017realtime}. Following skeleton estimation, two detection passes are performed to estimate heatmaps for keypoints of interest. The former of these acts as a coarse detector (5b in Figure~\ref{fig:efficientpose}), whereas the latter (5c in Figure~\ref{fig:efficientpose}) refines localization to yield more accurate outputs.

Note that in OpenPose, the heatmaps of the final detection pass are constrained to a low spatial resolution, which are incapable of achieving the amount of details that are normally inherent in the high-resolution input~\cite{cao2018openpose}. To improve this limitation of OpenPose, a series of three transposed convolutions performing bilinear upsampling are added for $8\times$ upscaling of the low-resolution heatmaps (step 6 in Figure~\ref{fig:openpose}). Thus, we project the low-resolution output onto a space of higher resolution in order to allow an increased level of detail. To achieve the proper level of interpolation while operating efficiently, each transposed convolution increases the map size by a factor of $2$, using a stride of $2$ with a $4 \times 4$ kernel.

\subsection{Variants}
\label{sec:variants}

Following the same principle as suggested in the original EfficientNet~\cite{tan2019efficientnet}, we scale the EfficientPose network architecture by adjusting the three main dimensions, i.e., input resolution, network width, and network depth, using the coefficients of Equation~\ref{eq:coefficients}. The results from this scaling are five different architecture variants that are given in Table~\ref{tab:variants}, referred to as EfficientPose I to IV and RT). As can be observed in this table, the input resolution, defined by the spatial dimensions of the image ($H \times W$), is scaled utilizing the high and low-level EfficientNet backbones that best match the resolution of high and low-resolution inputs (see Table~\ref{tab:efficientnets}). Here, the network width refers to the number of feature maps that are generated by each $E{\text -}MBConv6$. As described in Section~\ref{sec:architecture}, width scaling is achieved using the same width as the high-level backbone (i.e., $C$). The scaling of network depth is achieved in the number of Mobile DenseNets (i.e., $MD (C)$ in Table~\ref{tab:variants}) in the detection blocks. Also, this ensures that receptive fields across different models and spatial resolutions have similar relative sizes. For each model variant, we select the number ($D$) of Mobile DenseNets that best approximates the original depth factor $\alpha^{\phi}$ in the high-level EfficientNet backbone (Table~\ref{tab:efficientnets}). More specifically, the number of Mobile DenseNets are determined by Equation~\ref{eq:depth}, rounding to the closest integer. In addition to EfficientPose I to IV, the single-resolution model EfficientPose RT is formed to match the scale of the smallest EfficientNet model, providing HPE in extremely low latency applications.

\begin{equation}
D=\floor*{\alpha^{\phi}+0.5}
\label{eq:depth}
\end{equation}

\begin{table*}
\centering
\caption{Variants of EfficientPose obtained by scaling resolution, width, and depth. Mobile DenseNets $MD (C)$ computes $3C$ feature maps. $P$ and $Q$ denotes the number of 2D part affinity fields and confidence maps, respectively. $Conv^{T} (K \times K, O, S)$ defines transposed convolutions with kernel size $K \times K$, output maps $O$, and stride $S$}
\label{tab:variants}       
\begin{tabular}{|c|c|c|c|c|c|}
\hline
\textbf{Stage}        & \textbf{EfficientPose RT}   & \textbf{EfficientPose I}    & \textbf{EfficientPose II}    & \textbf{EfficientPose III}   & \textbf{EfficientPose IV}    \\ \hline
High-resolution input & $224 \times 224$                   & $256 \times 256$                   & $368 \times 368$                    & $480 \times 480$                    & $600 \times 600$                    \\ \hline
High-level backbone   & B0 (Block 1-3) & B2 (Block 1-3) & B4 (Block 1-3)  & B5 (Block 1-3)  & B7 (Block 1-3)  \\ \hline
Low-resolution input  & $-$                           & $128 \times 128$                   & $184 \times 184$                    & $240 \times 240$                    & $300 \times 300$                    \\ \hline
Low-level backbone    & $-$                           & B0 (Block 1-2) & B0 (Block 1-2)  & B1 (Block 1-2)  & B3 (Block 1-2)  \\ \hline
Detection block       & $MD (40)$        & $MD (48)$        & $[MD (56)] \times 2$ & $[MD (64)] \times 3$ & $[MD (80)] \times 4$ \\ \hline
Prediction pass 1     & \multicolumn{5}{c|}{$Conv (1 \times 1, 2P, 1)$}                                                                                                                 \\ \hline
Prediction pass 2-3   & \multicolumn{5}{c|}{$Conv (1 \times 1, Q, 1)$}                                                                                                                  \\ \hline
Upscaling             & \multicolumn{5}{c|}{$[Conv^{T} (4 \times 4, Q, 2)] \times 3$}                                                                                                         \\ \hline
\end{tabular}
\end{table*}

\subsection{Summary of proposed framework}
\label{sec:summary}

As can be inferred from the discussion above, the EfficientPose framework comprises a family of five ConvNets (i.e., EfficientPose I-IV and RT) that are constructed by compound scaling~\cite{tan2019efficientnet}. With this, EfficientPose exploits the advances in computationally efficient ConvNets for image recognition to construct a scalable network architecture that is capable of performing single-person HPE across different computational constraints. More specifically, EfficientPose utilizes both high and low-resolution images to provide two separate viewpoints that are processed independently through high and low-level backbones, respectively. The resulting features are concatenated to produce cross-resolution features, enabling selective emphasis on global and local image information. The detection stage employs a scalable mobile detection block to perform detection in three passes. The first pass estimates person skeletons through part affinity fields~\cite{cao2017realtime} to yield feasible pose configurations. The second and third passes estimate keypoint locations with progressive improvement in precision. Finally, the low-resolution prediction of the third pass is scaled up through bilinear interpolation to further improve the precision level.

\section{Experiments and results}
\label{sec:experiments}

\subsection{Experimental setup}
\label{sec:setup}

\begin{figure*}
\begin{center}
\includegraphics[width=\textwidth]{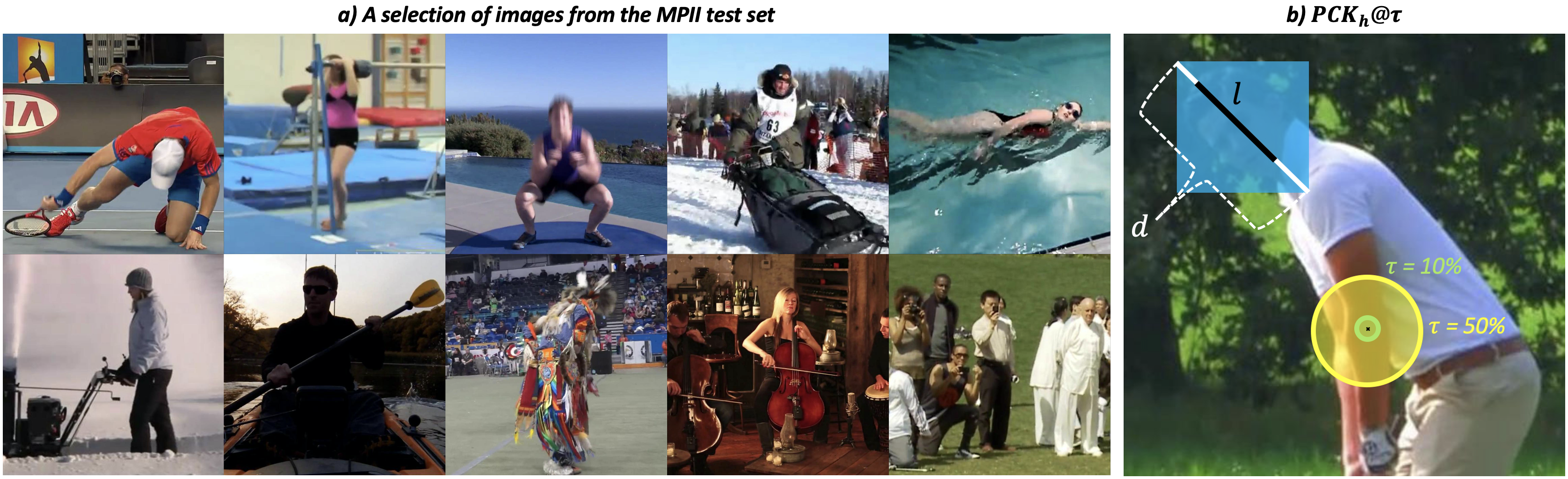}
\caption{The MPII single-person pose estimation challenge. From left: a) 10 images from the MPII test set displaying some of the variation and difficulties inherent in this challenge. b) The evaluation metrics $PCK_h@50$ and $PCK_h@10$ define the average of predictions within $\tau l$ distance ($l = 0.6d$) from the ground-truth location (e.g., left elbow), with $\tau$ being 50\% and 10\%, respectively}
\label{fig:mpii}     
\end{center}
\end{figure*}

We evaluate EfficientPose and compare it with OpenPose on the single-person MPII dataset~\cite{andriluka14cvpr}, containing images of mainly healthy adults in a wide range of different outdoor and indoor everyday activities and situations, such as sports, fitness exercises, housekeeping activities, and public events (Figure~\ref{fig:mpii}a). All models are optimized on MPII using stochastic gradient descent (SGD) on the mean squared error (MSE) of the model predictions relative to the target coordinates. More specifically, we applied SGD with momentum and cyclical learning rates (see Appendix~\ref{sec:optimization} for more information and further details on the optimization procedure). The learning rate is bounded according to the model-specific value of which it does not diverge during the first cycle ($\lambda_{max}$) and $\lambda_{min}=\frac{\lambda_{max}}{3000}$. The model backbones (i.e., VGG-19 for OpenPose, and EfficientNets for EfficientPose) are initialized with pretrained ImageNet weights, whereas the remaining layers employ random weight initialization. Supported by our experiments on training efficiency (see Appendix~\ref{sec:ablation}), we train the models for 200 epochs, except for OpenPose, which requires a higher number of epochs to converge (see Figure~\ref{fig:convergence} and Table~\ref{tab:epochs}). 

The training and validation portion of the dataset comprises $29$K images, and by adopting a standard random split, we obtain $26$K and $3$K instances for training and validation, respectively. We augment the images during training using random horizontal flipping, scaling ($0.75-1.25$), and rotation ($+$/$-$ $45$ degrees). We utilize a batch size of $20$, except for the high-resolutional EfficientPose III and IV, which both require a smaller batch size to fit into the GPU memory, $10$ and $5$, respectively. The experiments are carried out on an NVIDIA Tesla V100 GPU. 

The evaluation of model accuracy is performed using the $PCK_h@\tau$ metric. $PCK_h@\tau$ is defined as the fraction of predictions residing within a distance $\tau l$ from the ground truth location (see Figure~\ref{fig:mpii}b). $l$ is $60\%$ of the diagonal $d$ of the head bounding box, and $\tau$ the accepted percentage of misjudgment relative to $l$. $PCK_h@50$ is the standard performance metric for MPII but we also include the stricter $PCK_h@10$ metric for assessing models’ ability to yield highly precise keypoint estimates. As commonly done in the field, the final model predictions are obtained by applying multi-scale testing procedure~\cite{yang2017learning, sun2019deep, tang2018deeply}. Due to the restriction in the number of attempts for official evaluation on MPII, we only used the test metrics on the OpenPose baseline, and the most efficient and most accurate models, EfficientPose RT and EfficientPose IV, respectively. To measure model efficiency, both FLOPs and number of parameters are supplied.

\subsection{Results}
\label{sec:results}

Table~\ref{tab:mpiival} shows the results of our experiments with OpenPose and EfficientPose on the MPII validation dataset. As can be observed in this table, EfficientPose consistently outperformed OpenPose with regards to efficiency, with $2.2-184 \times$ reduction in FLOPs and $4-56 \times$ fewer number of parameters. In addition to this, all the model variants of EfficientPose achieved better high-precision localization, with a $0.8-12.9\%$ gain in \\ $PCK_h@10$ as compared to OpenPose. In terms of\\$PCK_h@50$, the high-end models, i.e., EfficientPose II-IV, managed to gain $0.6-2.2\%$ improvements against OpenPose. As Table~\ref{tab:mpiitest} depicts, EfficientPose IV achieved state-of-the-art results (a mean $PCK_h@50$ of 91.2) on the official MPII test dataset for models with number of parameters of a size less than $10$ million.

\begin{table*}
\centering
\caption{Performance of EfficientPose compared to OpenPose on the MPII validation dataset, as evaluated by efficiency (number of parameters and FLOPs, and relative reduction in parameters and FLOPs compared to OpenPose) and accuracy (mean $PCK_h@50$ and mean $PCK_h@10$)}
\label{tab:mpiival}       
\begin{tabular}{lllllll}
\hline\noalign{\smallskip}
\textbf{Model} & \textbf{Parameters} & \textbf{Parameter reduction} & \textbf{FLOPs} & \textbf{FLOP reduction} & $\mathbf{PCK_h@50}$ & $\mathbf{PCK_h@10}$  \\
\noalign{\smallskip}\hline\noalign{\smallskip}
OpenPose~\cite{cao2018openpose} & $25.94M$ &	$1\times$ & $160.36G$ & $1\times$ & $87.60$	& $22.76$ \\
EfficientPose RT & $0.46M$ & $56\times$ & $0.87G$ & $184\times$ & $82.88$ & $23.56$ \\
EfficientPose I	& $0.72M$ & $36\times$ & $1.67G$ & $96\times$ & $85.18$ & $26.49$ \\
EfficientPose II & $1.73M$ & $15\times$ & $7.70G$ & $21\times$ & $88.18$ & $30.17$ \\
EfficientPose III & $3.23M$ & $8.0\times$ & $23.35G$ & $6.9\times$ & $89.51$ & $30.90$ \\
EfficientPose IV & $6.56M$ & $4.0\times$ & $72.89G$ & $2.2\times$ & $89.75$ & $35.63$ \\
\noalign{\smallskip}\hline
\end{tabular}
\end{table*}

\begin{table*}
\centering
\caption{State-of-the-art results in $PCK_h@50$ (both for individual body parts and overall mean value) on the official MPII test dataset~\cite{andriluka14cvpr} compared to the number of parameters}
\label{tab:mpiitest}       
\begin{tabular}{llllllllll}
\hline\noalign{\smallskip}
\textbf{Model} & \textbf{Parameters} & \textbf{Head} & \textbf{Shoulder} & \textbf{Elbow} & \textbf{Wrist} & \textbf{Hip} & \textbf{Knee} & \textbf{Ankle} & \textbf{Mean}  \\
\noalign{\smallskip}\hline\noalign{\smallskip}
Pishchulin et al., ICCV’13~\cite{pishchulin2013poselet} & $-$ & $74.3$ & $49.0$ & $40.8$ & $32.1$ & $36.5$ & $34.4$ & $35.2$ & $44.1$ \\
Tompson et al., NIPS’14~\cite{tompson2014joint} & $-$ & $95.8$ & $90.3$ & $80.5$ & $74.3$ & $77.6$ & $69.7$ & $62.8$ & $79.6$ \\
Lifshitz et al., ECCV’16~\cite{lifshitz2016human} & $76M$ & $97.8$ & $93.3$ & $85.7$ & $80.4$ & $85.3$ & $76.6$ & $70.2$ & $85.0$ \\
Tang et al., BMVC'18~\cite{tang2018cu} & $10M$ & $97.4$ & $96.2$ & $91.8$ & $87.3$ & $90.0$ & $87.0$ & $83.3$ & $90.8$ \\
Newell et al., ECCV’16~\cite{newell2016stacked} & $26M$ & $98.2$ & $96.3$ & $91.2$ & $87.1$ & $90.1$ & $87.4$ & $83.6$ & $90.9$ \\
Zhang et al., CVPR’19~\cite{zhang2019fast} & $3M$ & $98.3$ & $96.4$ & $91.5$ & $87.4$ & $90.9$ & $87.1$ & $83.7$ & $91.1$ \\
Bulat et al., FG'20~\cite{bulat2020toward} & $9M$ & $98.5$ & $96.4$ & $91.5$ & $87.2$ & $90.7$ & $86.9$ & $83.6$ & $91.1$ \\
Yang et al., ICCV’17~\cite{yang2017learning} & $27M$ & $98.5$ & $96.7$ & $92.5$ & $88.7$ & $91.1$ & $88.6$ & $86.0$ & $92.0$ \\
Tang et al., ECCV’18~\cite{tang2018deeply} & $16M$ & $98.4$ & $96.9$ & $92.6$ & $88.7$ & $91.8$ & $89.4$ & $86.2$ & $92.3$ \\
Sun et al., CVPR’19~\cite{sun2019deep} & $29M$ & $98.6$ & $96.9$ & $92.8$ & $89.0$ & $91.5$ & $89.0$ & $85.7$ & $92.3$ \\
Zhang et al., arXiv’19~\cite{zhang2019human} & $24M$ & $98.6$ & $97.0$ & $92.8$ & $88.8$ & $91.7$ & $89.8$ & $86.6$ & $92.5$ \\
\noalign{\smallskip}\hline\noalign{\smallskip}
OpenPose~\cite{cao2018openpose} & $25.94M$ & $97.7$ & $94.7$ & $89.5$ & $84.7$ & $88.4$ & $83.6$ & $79.3$ & $88.8$ \\
EfficientPose RT & $0.46M$ & $97.0$ & $93.3$ & $85.0$ & $79.2$ & $85.9$ & $77.0$ & $71.0$ & $84.8$ \\
EfficientPose IV & $6.56M$ & $98.2$ & $96.0$ & $91.7$ & $87.9$ & $90.3$ & $87.5$ & $83.9$ & $91.2$ \\
\noalign{\smallskip}\hline
\end{tabular}
\end{table*}

\begin{figure*}
\begin{center}
\includegraphics[width=\textwidth]{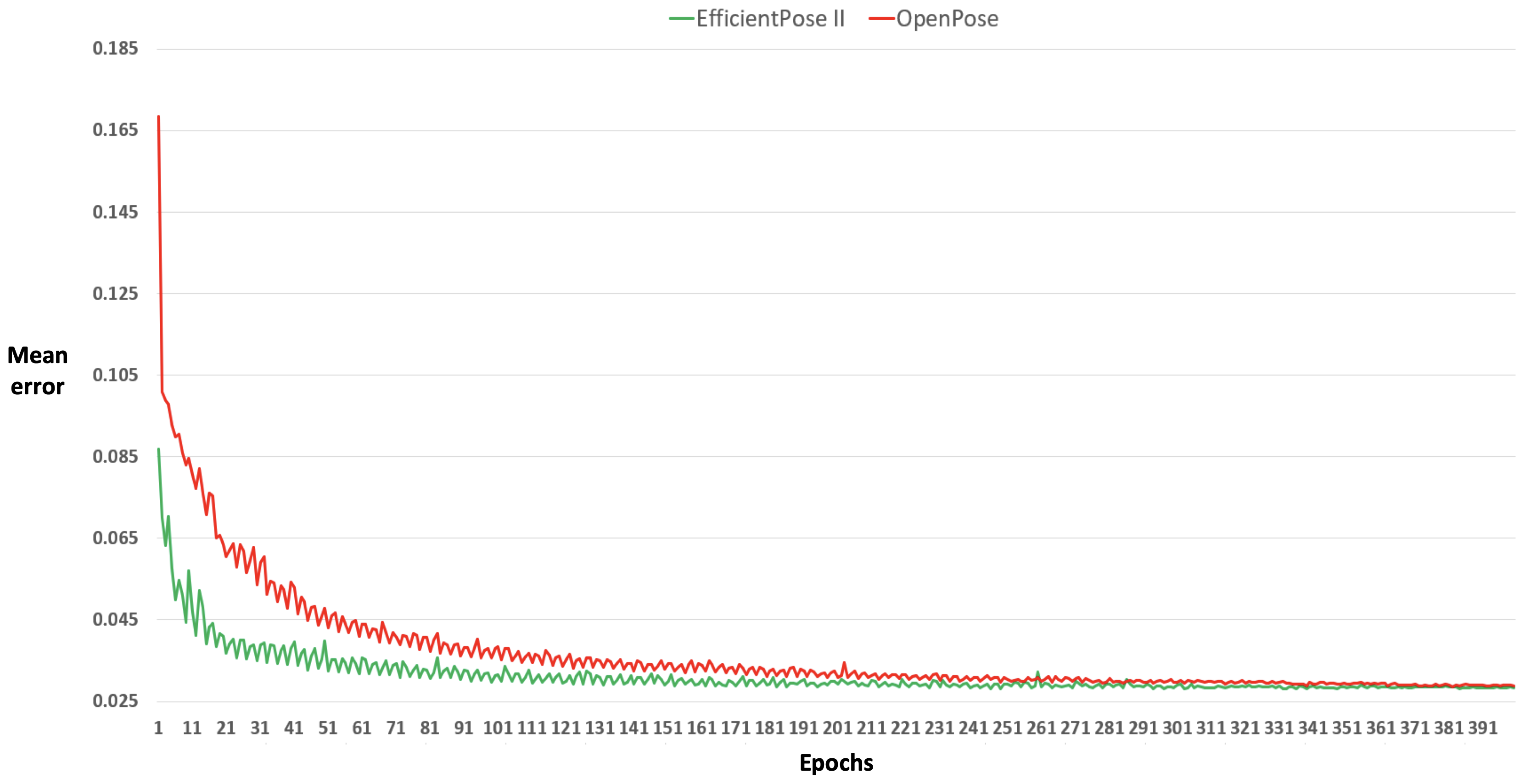}
\caption{The progression of the mean error of EfficientPose II and OpenPose on the MPII validation set during the course of training}
\label{fig:convergence}       
\end{center}
\end{figure*}

Compared to OpenPose, EffcientPose also exhibited rapid convergence during training. We optimized both approaches on similar input resolution, which defaults to $368 \times 368$ for OpenPose, corresponding to EfficientPose II. The training graph shown in Figure~\ref{fig:convergence} demonstrates that EfficientPose converges early, whereas OpenPose requires up to $400$ epochs before achieving proper convergence. Nevertheless, OpenPose benefited from this prolonged training in terms of precision, with a $2.6\%$ improvement in $PCK_h@50$ during the final $200$ epochs, whereas EfficientPose II had a minor gain of $0.4\%$ (see Table~\ref{tab:epochs}).

\begin{table*}
\centering
\caption{Model accuracy on the MPII validation dataset in relation to the number of training epochs}
\label{tab:epochs}       
\begin{tabular}{lll}
\hline\noalign{\smallskip}
\textbf{Model} & \textbf{Epochs} & $\mathbf{PCK_h@50}$  \\
\noalign{\smallskip}\hline\noalign{\smallskip}
OpenPose~\cite{cao2018openpose} & $100$ & $80.47$ \\
OpenPose~\cite{cao2018openpose} & $200$ & $85.00$ \\
OpenPose~\cite{cao2018openpose} & $400$ & $87.60$ \\
EfficientPose II & $100$ & $87.05$ \\
EfficientPose II & $200$ & $88.18$ \\
EfficientPose II & $400$ & $88.56$ \\
\noalign{\smallskip}\hline
\end{tabular}
\end{table*}

\section{Discussion}
\label{sec:discussion}

In this section, we discuss several aspects of our findings and possible avenues for further research.

\subsection{Improvements over OpenPose}
\label{sec:improvement}

The precision of HPE methods is a key success factor for analyses of movement kinematics, like segment positions and joint angles, for assessment of sport performance in athletes, or motor disabilities in patients. Facilitated by cross-resolution features and upscaling of output (see Appendix~\ref{sec:ablation}), EfficientPose achieved a higher precision than OpenPose~\cite{cao2018openpose}, with a $57\%$ relative improvement in $PCK_h@10$ on single-person MPII (Table~\ref{tab:mpiival}). What this means is that the EfficientPose architecture is generally more suitable in performing precision-demanding single-person HPE applications, like medical assessments and elite sports, than OpenPose. 

Another aspect to have in mind is that, for some applications (e.g., exercise games and baby monitors), we might be more interested in the latency of the system and its ability to respond quickly. Hence, the degree of correctness in keypoint predictions might be less crucial. In such scenarios, with applications that demand high-speed predictions, the 460K parameter model, EfficientPose RT, consuming less than one GFLOP, would be suitable. Nevertheless, it still manages to provide higher precision level than current approaches in the high-speed regime, e.g., \cite{bulat2020toward, tang2018cu}. Further, the scalability of EfficientPose enables flexibility in various situations and across different types of hardware, whereas OpenPose suffers from its large number of parameters and computational costs (FLOPs). 

\subsection{Strengths of the EfficientPose approach}
\label{sec:strengths}

The use of MBConv in HPE is to the best of our knowledge an unexplored research area. This has also been partly our main motivation for exploring the use of MBConv in our EfficientPose approach, recognizing its success in image classification~\cite{tan2019efficientnet}. Our experimental results showed that EfficientPose approached state-of-the-art performance on the single-person MPII benchmark despite a large reduction in the number of parameters (Table~\ref{tab:mpiitest}). This means that the parameter-efficient MBConvs provide value in HPE as with other computer vision tasks, such as image classification and object detection. This, in turns, makes MBConv a very suitable component for HPE networks. For this reason, it would be interesting to investigate the effect of combining it with other novel HPE architectures, such as Hourglass and HRNet~\cite{newell2016stacked, sun2019deep}. 

Further, the use of EfficientNet as a backbone, and the proposed cross-resolution feature extractor combining several EfficientNets for improved handling of basic features, are also interesting avenues to explore further. From the present study, it is reasonable to assume that EfficientNets could replace commonly used backbones for HPE, such as VGG and ResNets, which would reduce the computational overheads associated with these approaches~\cite{simonyan2014very, he2016deep}. Also, a cross-resolution feature extractor could be useful for precision-demanding applications by providing an improved performance on $PCK_h@10$ (Table~\ref{tab:features}). 

We also observed that EfficientPose benefited from compound model scaling across resolution, width and depth. This benefit was reflected by the increasing improvements in $PCK_h@50$ and $PCK_h@10$ from EfficientPose RT through EfficientPose I to EfficientPose IV (Table~\ref{tab:mpiival}). To conclude, we can exploit this to further examine scalable ConvNets for HPE, and thus obtain insights into appropriate sizes of HPE models (i.e., number of parameters), required number of FLOPs, and obtainable precision levels.

In this study, OpenPose and EfficientPose were optimized on the general-purpose MPII Human Pose Dataset. For many applications (e.g., action recognition and video surveillance) the variability in MPII may be sufficient for directly applying the models on real-world problems. Nonetheless, there are other particular scenarios that deviate from the setting addressed in this paper. The MPII dataset comprises mostly healthy adults in a variety of every day indoor and outdoor activities~\cite{andriluka14cvpr}. In less natural environments (e.g., movement science laboratories or hospital settings) and with humans of different anatomical proportions such as children and infants~\cite{sciortino2017estimation}, careful consideration must be taken. This could include a need for fine-tuning of the MPII models on more specific datasets related to the problem at hand. As mentioned earlier, our experiments showed that EfficientPose was more easily trainable than OpenPose (Figure~\ref{fig:convergence} and Table~\ref{tab:epochs}). This trait of rapid convergence suggests that exploring the use of transfer learning on the EfficientPose models on other HPE data could provide interesting results. 

\subsection{Avenues for further research}
\label{sec:avenues}

The precision level of pose configurations provided by EfficientPose in the context of target applications is a topic considered beyond the scope of this paper and has for this reason been left for further studies. We can establish the validity of EfficientPose for robust single-person pose estimation already by examining whether the movement information supplied by the proposed framework is of sufficiently good quality for tackling challenging problems, such as complex human behavior recognition~\cite{liu2018learning, fernando2018tracking}. To assess this, we could, for example, compare the precision level of the keypoint estimates supplied by EfficientPose with the movement information provided by body-worn movement sensors. Moreover, we could combine the proposed image-based EfficientPose models with body-worn sensors, such as inertial measurement unit (IMU)~\cite{kundu2018hand}, or physiological signals, like electrical cardiac activity and electrical brain activity~\cite{fiorini2020unsupervised}, to potentially achieve improved precision levels and an increased robustness. Our hypothesis is that using body-worn sensors or physiological instruments could be useful in situations where body parts are extensively occluded, such that camera-based recognition alone may not be sufficient for accurate pose estimation.

Another path for further study and validation is the capability of EfficientPose to perform multi-person HPE. The improved computational efficiency of EfficientPose compared to OpenPose has the potential to also benefit multi-person HPE. State-of-the-art methods for multi-person HPE are dominated by top-down approaches, which require computation that is normally proportional to the number of individuals in the image~\cite{fieraru2018learning, zhang2019distribution}. In crowded scenes, top-down approaches are highly resource demanding. Similar to the original OpenPose~\cite{cao2018openpose}, and few other recent works on multi-person HPE~\cite{huang2020high, guan2019realtime}, EfficientPose incorporates part affinity fields, which would enable the grouping of keypoints into persons, and thus allowing to perform multi-person HPE in a bottom-up manner. This would reduce the computational overhead into a single network inference per image, and hence yield more computationally efficient multi-person HPE.

Further, it would be interesting to explore the extension of the proposed framework to perform 3D pose estimation as part of our future research. In accordance with recent studies, 3D pose projection from 2D images can be achieved, either by employing geometric relationships between 2D keypoint positions and 3D human pose models~\cite{yuan2020single}, or by leveraging occlusion-robust pose-maps (ORPM) in combination with annotated 3D poses~\cite{mehta2018single, benzine2020single}.

The architecture of EfficientPose and the training process can be improved in several ways. First, the optimization procedure (see Appendix~\ref{sec:optimization}) was developed for maximum $PCK_h@50$ accuracy on OpenPose, and simply reapplied to EfficientPose. Other optimization procedures might be more appropriate, including alternative optimizers (e.g., Adam~\cite{kingma2014adam} and RMSProp~\cite{tieleman2012lecture}), and other learning rate and sigma schedules. 

Second, only the backbone of EfficientPose was pretrained on ImageNet. This could restrict the level of accuracy on HPE because large-scale pretraining not only supplies robust basic features but also higher-level semantics. Thus, it would be valuable to assess the effect of pretraining on model precision in HPE. We could, for example, pretrain the majority of ConvNet layers on ImageNet, and retrain these on HPE data. 

Third, the proposed compound scaling of EfficientPose assumes that the scaling relationship between resolution, width, and depth, as defined by Equation~\ref{eq:coefficients}, is identical in HPE and image classification. However, the optimal compound scaling coefficients might be different for HPE, where the precision level is more dependent on image resolution, than for image classification. Based on this, a topic for further studies could be to conduct neural architecture search across different combinations of resolution, width, and depth in order to determine the optimal combination of scaling coefficients for HPE. Regardless of the scaling coefficients, the scaling of detection blocks in EfficientPose could be improved. The block depth (i.e., number of Mobile DenseNets) slightly deviates from the original depth coefficient in EfficientNets based on the rigid nature of the Mobile DenseNets. A carefully designed detection block could address this challenge by providing more flexibility with regards to the number of layers and the receptive field size.

Fourth, the computational efficiency of EfficientPose could be further improved by the use of teacher-student network training (i.e., knowledge distillation)~\cite{bucilua2006model} to transfer knowledge from a high-scale EfficientPose teacher network to a low-scale EfficientPose student network. This technique has already shown promising results in HPE when paired with the stacked hourglass architecture~\cite{newell2016stacked, zhang2019fast}. Sparse networks, network pruning, and weight quantization~\cite{tung2018clip, elsen2019fast} could also be included in the study to facilitate the development of more accurate and responsive real-life systems for HPE. Finally, for high performance inference and deployment on edge devices, further speed-up could be achieved by the use of specialized libraries such as NVIDIA TensorRT and TensorFlow Lite~\cite{TensorRT:2020, TensorFlowLite:2020}.

In summary, EfficientPose tackles single-person HPE with an improved degree of precision compared to the commonly adopted OpenPose network~\cite{cao2018openpose}. In addition to this, the EfficientPose models have the ability to yield high performance with a large reduction in number of parameters and FLOPs. This has been achieved by exploiting the findings from contemporary research within image recognition on computationally efficient ConvNet components, most notably MBConvs and EfficientNets~\cite{sandler2018mobilenetv2, tan2019efficientnet}. Again, for the sake of reproducibility, we have made the EfficientPose models publicly available for other researchers to test and possibly further development.

\section{Conclusion}
\label{sec:conclusion}

In this work, we have stressed the need for a publicly accessible method for single-person HPE that suits the demands for both precision and efficiency across various applications and computational budgets. To this end, we have presented a novel method called EfficientPose, which is a scalable ConvNet architecture leveraging a computationally efficient multi-scale feature extractor, novel mobile detection blocks, skeleton estimation, and bilinear upscaling. In order to have model variants that are able to flexibly find a sensible trade-off between accuracy and efficiency, we have exploited model scalability in three dimensions: input resolution, network width, and network depth. Our experimental results have demonstrated that the proposed approach has the capability to offer computationally efficient models, allowing real-time inference on edge devices. At the same time, our framework offers flexibility to be scaled up to deliver more precise keypoint estimates than commonly used counterparts, at an order of magnitude less parameters and computational costs (FLOPs). Taking into account the efficiency and high precision level of our proposed framework, there is a reason to believe that EfficientPose will provide an important foundation for the next-generation markerless movement analysis. 

In our future work, we plan to develop new techniques to further improve the model effectiveness, especially in terms of precision, by investigating optimal compound model scaling for HPE. Moreover, we will deploy EfficientPose on a range of applications to validate its applicability, as well as feasibility, in real-world scenarios.

\section*{Acknowledgements}
\label{sec:acknowledgements}

The research is funded by RSO funds from the Faculty of Medicine and Health Sciences at the Norwegian University of Science and Technology. The experiments were carried out utilizing computational resources provided by the Norwegian Open AI Lab.

%
%

\bibliographystyle{spmpsci}      
\bibliography{references} 

\appendix
\section*{Appendices}
\section{Ablation study}
\label{sec:ablation}

To determine the effect of different design choices in the EfficientPose architecture, we carried out component analysis. 

\subsection*{Training efficiency}
\label{sec:training}

We assessed the number of training epochs to determine the appropriate duration of training, avoiding demanding optimization processes. Figure~\ref{fig:convergence} suggests that the largest improvement in model accuracy occurs until around $200$ epochs, after which training saturates. Table~\ref{tab:epochs} supports this observation with less than $0.4\%$ increase in $PCK_h@50$ with $400$ epochs of training. From this, it was decided to perform the final optimization of the different variants of EfficientPose over $200$ epochs. Table~\ref{tab:epochs} also suggests that most of the learning progress occurs during the first $100$ epochs. Hence, for the remainder of the ablation study $100$ epochs were used to determine the effect of different design choices.

\subsection*{Cross-resolution features}
\label{sec:features}

The value of combining low-level local information with high-level semantic information through a cross-resolution feature extractor was evaluated by optimizing the model with and without the low-level backbone. Experiments were conducted on two different variants of the EfficientPose model. On coarse prediction ($PCK_h@50$) there is little to no gain in accuracy (Table~\ref{tab:features}), whereas for fine estimation ($PCK_h@10$) some improvement ($0.6-0.7\%$) is displayed taking into account the negligible cost of $1.02-1.06 \times$ more parameters and $1.03-1.06 \times$ increase in FLOPs. 

\begin{table*}
\centering
\caption{Model accuracy on the MPII validation dataset in relation to the use of cross-resolution features}
\label{tab:features}       
\begin{tabular}{llllll}
\hline\noalign{\smallskip}
\textbf{Model} & \textbf{Cross-resolution features} & \textbf{Parameters} & \textbf{FLOPs} & $\mathbf{PCK_h@50}$ & $\mathbf{PCK_h@10}$  \\
\noalign{\smallskip}\hline\noalign{\smallskip}
EfficientPose I & \checkmark & $0.72M$ & $1.67G$ & $83.56$ & $26.35$ \\
EfficientPose I &  & $0.68M$ & $1.58G$ & $83.64$ & $25.79$ \\
EfficientPose II & \checkmark & $1.73M$ & $7.70G$ & $87.05$ & $29.87$ \\
EfficientPose II & & $1.69M$ & $7.50G$ & $86.93$ & $29.16$ \\
\noalign{\smallskip}\hline
\end{tabular}
\end{table*}

\subsection*{Skeleton estimation}
\label{sec:skeleton}

The effect of skeleton estimation through the approximation of part affinity fields was assessed by comparing the architecture with and without the single pass of skeleton estimation. Skeleton estimation yields improved accuracy with $1.3-2.4\%$ gain in $PCK_h@50$ and $0.2-1.4\%$ in $PCK_h@10$ (Table~\ref{tab:skeleton}), while only introducing an overhead in number of parameters and computational cost of $1.3-1.4 \times$ and $1.2-1.3 \times$, respectively.

\begin{table*}
\centering
\caption{Model accuracy on the MPII validation dataset in relation to the use of skeleton estimation}
\label{tab:skeleton}       
\begin{tabular}{llllll}
\hline\noalign{\smallskip}
\textbf{Model} & \textbf{Skeleton estimation} & \textbf{Parameters} & \textbf{FLOPs} & $\mathbf{PCK_h@50}$ & $\mathbf{PCK_h@10}$  \\
\noalign{\smallskip}\hline\noalign{\smallskip}
EfficientPose I & \checkmark & $0.72M$ & $1.67G$ & $83.56$ & $26.35$ \\
EfficientPose I &  & $0.54M$ & $1.37G$ & $81.13$ & $25.00$ \\
EfficientPose II & \checkmark & $1.73M$ & $7.70G$ & $87.05$ & $29.87$ \\
EfficientPose II & & $1.27M$ & $6.03G$ & $85.75$ & $29.67$ \\
\noalign{\smallskip}\hline
\end{tabular}
\end{table*}

\subsection*{Number of detection passes}
\label{sec:passes}

We also determined the appropriate comprehensiveness of detection, represented by number of detection passes. EfficientPose I and II were both optimized on three different variants (Table~\ref{tab:passes}). Seemingly, the models benefit from intermediate supervision with a general trend of increased performance level in accordance with number of detection passes. The major benefit in performance is obtained by expanding from one to two passes of keypoint estimation, reflected by $1.6-1.7\%$ increase in $PCK_h@50$ and $1.8-1.9\%$ in $PCK_h@10$. In comparison, a third detection pass yields only $0.5-0.8\%$ relative improvement in $PCK_h@50$ compared to two passes, and no gain in $PCK_h@10$ while increasing number of parameters and computation by $1.3 \times$ and $1.2 \times$, respectively. From these findings, we decided a beneficial trade-off in accuracy and efficiency would be the use of two detection passes.

\begin{table*}
\centering
\caption{Model accuracy on the MPII validation dataset in relation to the number of detection passes}
\label{tab:passes}       
\begin{tabular}{llllll}
\hline\noalign{\smallskip}
\textbf{Model} & \textbf{Detection passes} & \textbf{Parameters} & \textbf{FLOPs} & $\mathbf{PCK_h@50}$ & $\mathbf{PCK_h@10}$  \\
\noalign{\smallskip}\hline\noalign{\smallskip}
EfficientPose I & $1$ & $0.52M$ & $1.33G$ & $81.85$ & $24.51$ \\
EfficientPose I & $2$ & $0.72M$ & $1.67G$ & $83.56$ & $26.35$ \\
EfficientPose I & $3$ & $0.92M$ & $2.02G$ & $84.35$ & $26.42$ \\
EfficientPose II & $1$ & $1.24M$ & $5.92G$ & $85.42$ & $28.01$ \\
EfficientPose II & $2$ & $1.73M$ & $7.70G$ & $87.05$ & $29.87$ \\
EfficientPose II & $3$ & $2.22M$ & $9.49G$ & $87.55$ & $29.61$ \\
\noalign{\smallskip}\hline
\end{tabular}
\end{table*}

\subsection*{Upscaling}
\label{sec:upscaling}

To assess the impact of upscaling, implemented as bilinear transposed convolutions, we compared the results of the two respective models. Table~\ref{tab:upscaling} reflects that upscaling yields improved precision on keypoint estimates by large gains of $9.2-12.3\%$ in $PCK_h@10$ and smaller improvements of $0.5-1.1\%$ on coarse detection ($PCK_h@50$). As a consequence of increased output resolution upscaling slightly increases number of FLOPs ($1.04-1.1 \times$) with neglectable increase in number of parameters.

\begin{table*}
\centering
\caption{Model accuracy on the MPII validation dataset in relation to the use of upscaling}
\label{tab:upscaling}       
\begin{tabular}{llllll}
\hline\noalign{\smallskip}
\textbf{Model} & \textbf{Upscaling} & \textbf{Parameters} & \textbf{FLOPs} & $\mathbf{PCK_h@50}$ & $\mathbf{PCK_h@10}$  \\
\noalign{\smallskip}\hline\noalign{\smallskip}
EfficientPose I & \checkmark & $0.72M$ & $1.67G$ & $83.56$ & $26.35$ \\
EfficientPose I &  & $0.71M$ & $1.52G$ & $82.42$ & $14.02$ \\
EfficientPose II & \checkmark & $1.73M$ & $7.70G$ & $87.05$ & $29.87$ \\
EfficientPose II & & $1.73M$ & $7.37G$ & $86.56$ & $20.66$ \\
\noalign{\smallskip}\hline
\end{tabular}
\end{table*}

\section{Optimization procedure}
\label{sec:optimization}

Most state-of-the-art approaches for single-person pose estimation are extensively pretrained on ImageNet~\cite{sun2019deep, zhang2019human}, enabling rapid convergence for models when adapted to other tasks, such as HPE. In contrast to these approaches, few models, including OpenPose~\cite{cao2018openpose} and EfficientPose, only utilize the most basic pretrained features. This facilitates construction of more efficient network architectures but at the same time requires careful design of optimization procedures for convergence towards reasonable parameter values.

Training of pose estimation models is complicated due to the intricate nature of output responses. Overall, optimization is performed in a conventional fashion by minimizing the MSE of the predicted output maps $Y$ with respect to ground truth values $\hat{Y}$ across all output responses $N$.

The predicted output maps should ideally have higher values at the spatial locations corresponding to body part positions, while punishing predictions farther away from the correct location. As a result, the ground truth output maps must be carefully designed to enable proper convergence during training. We achieve this by progressively reducing the circumference from the true location that should be rewarded, defined by the $\sigma$ parameter. Higher probabilities $T \in [0,1]$ are assigned for positions $P$ closer to the ground truth position $G$ (Equation~\ref{eq:confidence}). 

\begin{equation}
T_i = \exp{(-\frac{\norm{P_i - G}_2^2}{\sigma^{2}})}
\label{eq:confidence}
\end{equation}

\begin{figure*}
\begin{center}
\includegraphics[width=\textwidth]{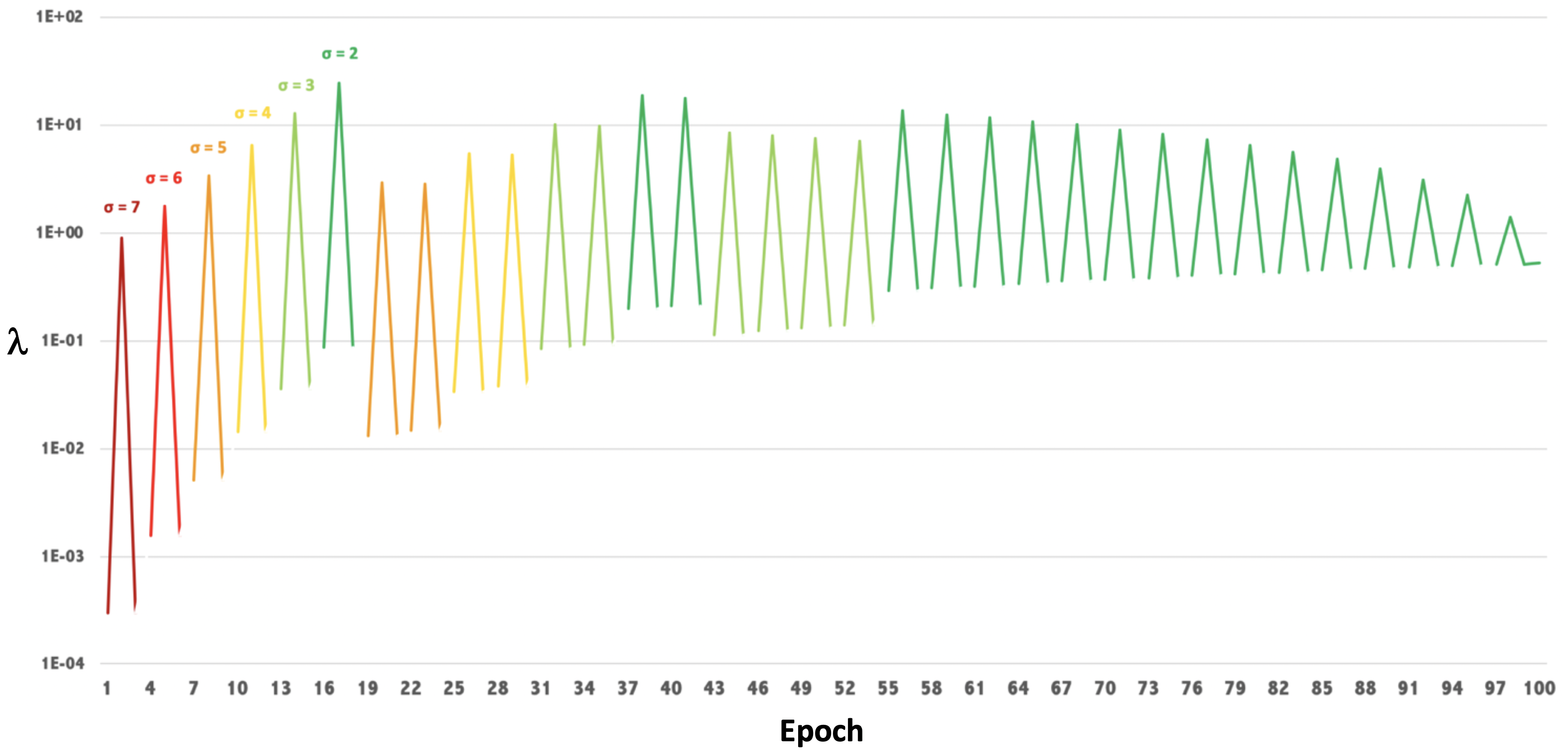}
\caption{Optimization scheme displaying learning rates $\lambda$ and $\sigma$ values corresponding to the training of EfficientPose II over $100$ epochs}
\label{fig:optimization}       
\end{center}
\end{figure*}

The proposed optimization scheme (Figure~\ref{fig:optimization}) incorporates a stepwise $\sigma$ scheme, and utilizes SGD with momentum of $0.9$ and a decaying triangular cyclical learning rate (CLR) policy~\cite{smith2017cyclical}. The $\sigma$ parameter is normalized according to the output resolution. As suggested by Smith and Topin~\cite{smith2019super}, the large learning rates in CLR provides regularization in network optimization. This makes training more stable and may even increase training efficiency. This is valuable for network architectures, such as OpenPose and EfficientPose, less heavily concerned with pretraining (i.e., having larger portions of randomized weights). In our adoption of CLR, we utilize a cycle length of $3$ epochs. The learning rate ($\lambda$) converges towards $\lambda_{\infty}$ (Equation~\ref{eq:lrfinal}), where $\lambda_{max}$ is the highest learning rate for which the model does not diverge during the first cycle and $\lambda_{min}=\frac{\lambda_{max}}{3000}$, whereas $\sigma_{0}$ and $\sigma_{\infty}$ are the initial and final sigma values, respectively.

\begin{equation}
\lambda_{\infty} = 10^{\frac{\log{(\lambda_{max})}+\log{(\lambda_{min})}}{2}} \cdot 2^{\sigma_{0}-\sigma_{\infty}}
\label{eq:lrfinal}
\end{equation}

\end{document}